\documentclass[letterpaper, 10 pt, conference]{ieeeconf}  

\IEEEoverridecommandlockouts                              

\usepackage{graphics} 
\usepackage{epsfig} 
\usepackage{graphicx}
\usepackage{subcaption}
\usepackage{amsmath} 
\usepackage{amssymb}  
\usepackage{xcolor}
\usepackage{multirow}
\usepackage{hyperref}
\usepackage{url}
\usepackage{array}
\usepackage{siunitx}

\usepackage{multirow}
\usepackage{xcolor}
\usepackage{pifont}   
\usepackage{array}    
\usepackage{cleveref}

\newcommand{\cmark}{\textcolor{green!60!black}{\ding{51}}}
\newcommand{\xmark}{\textcolor{red!70!black}{\ding{55}}}

\title{\LARGE \bf
Language-Guided Grasping under Partial Observation for Mobile Manipulation in Field Inspection and Maintenance
}

\author{Dilermando Almeida$^{1, \dagger}$, Juliano Negri$^{2, \dagger}$, Guilherme Lazzarini$^{2}$,
Thiago H. Segreto$^{2}$, Ranulfo Bezerra$^{3}$\\Gustavo J. G. Lahr$^{4}$, Ricardo V. Godoy$^{2,*}$ and Marcelo Becker$^{2}$
\thanks{
This work was supported by the Petr\'{o}leo Brasileiro S/A - Petrobras,
using resources from the R\&D clause of the ANP, in partnership with the
Universidade de S\~{a}o Paulo (USP) and the Funda\c{c}\~{a}o de Apoio \`{a} F\'{\i}sica e
\`{a} Qu\'{\i}mica (FAFQ), under Cooperation Agreement No. 2023/00016-6 and
2023/00013-7.; by the S\~ao Paulo Research Foundation~(FAPESP), Grant \#2025/08520-0.}
\thanks{$^{1}$Dilermando Almeida is with the Department of Mechanical Engineering, Federal University of Uberlândia, Uberlândia, Brazil}%
\thanks{$^{2}$Juliano D. Negri, Guilherme Lazzarini, Thiago H. Segreto, Ricardo V. Godoy, and Marcelo Becker are with the  Department of Mechanical Engineering, University of São Paulo, São Carlos, Brazil.}%
\thanks{$^{3}$Ranulfo Bezerra is with the Graduate School of Information Sciences, Tohoku University, Sendai, Japan.}%
\thanks{$^{4}$Gustavo is with the Faculdade Israelita de Ensino e Pesquisa Albert Einstein, Hospital Israelita Albert Einstein, Brazil.}%
\thanks{$^{\dagger}$These authors contributed equally to this work.}
\thanks{$^{*}$Corresponding author: {\tt\small ricardo.godoy@usp.br}}%
}

\begin{document}

\maketitle
\thispagestyle{empty}
\pagestyle{empty}

\begin{abstract}

Offshore inspection and maintenance have increasingly been using legged robots for routine sensing, yet many useful interventions still require physical interaction with tools, containers, and task-relevant objects. Employing robots for these tasks can reduce operators' exposure in confined, elevated, or potentially explosive areas. This paper presents a language-guided grasping pipeline for a legged mobile manipulator operating under partial observation. An operator defines the target, the system grounds it in RGB with open-vocabulary detection and promptable segmentation, extracts an object-centric RGB-D point cloud, improves sparse geometry through depth compensation and point-cloud completion, and selects a 6-DoF grasp using collision, clearance, reachability, and approach constraints. The system is implemented on a quadruped robot with an arm and evaluated in two cluttered tabletop scenes motivated by small-object retrieval during inspection and maintenance. Across paired trials, the proposed pipeline achieved 9/10 successful grasps, compared with 3/10 for a view-dependent deployment baseline. In this controlled setting, object-centric completion and execution-aware selection reduced approach collisions and improved the reliability of language-guided grasping for supervised field manipulation. Project page: \url{https://eesc-labrom.github.io/agn_grasp/}.

\end{abstract}

\section{INTRODUCTION}

Inspection and intervention in critical infrastructure, such as offshore oil and gas platforms, nuclear facilities, tunnels, and disaster-affected industrial sites, are central applications of safety, security, and rescue robotics. Legged platforms are useful in these environments because they can negotiate stairs, grating, narrow passages, and cluttered access routes while carrying visual, thermal, acoustic, or gas sensors~\cite{hutter2017anymal,gehring2021anymal,HALDER2023105814,9885187}. In such scenarios, manipulation is often required for safety-relevant intervention, such as retrieving loose objects, handling tools, opening panels, collecting samples, or removing obstacles without exposing human operators to hazardous areas. Moreover, many maintenance and intervention tasks still require physical contact with tools, small equipment, panels, samples, or loose objects~\cite{bengel2009mobile,hooft2024heights,galin2019automation}. For inspection and maintenance teams, for practical deployment, the robot should acquire a named object without colliding with the surrounding structure, using the imperfect viewpoint available at the work site.

\begin{figure}[!t]
    \centering
    \includegraphics[width=\linewidth]{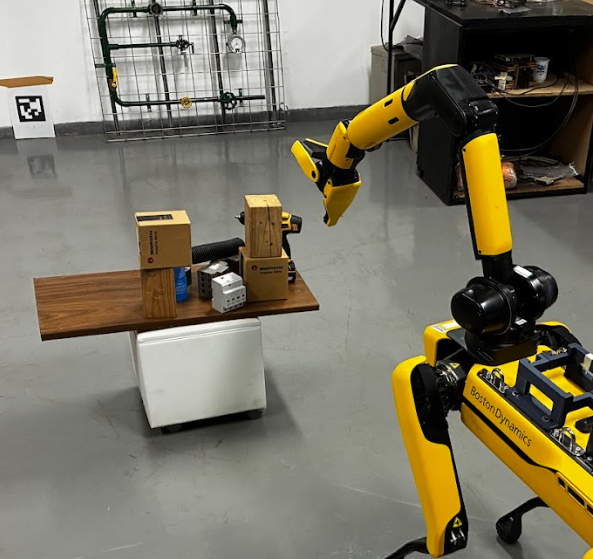}
    \caption{A legged mobile manipulator performs language-guided grasping in a cluttered scene under partial observation. The motivating use case is supervised manipulation during inspection and maintenance, where a remote operator may ask the robot to acquire a partially visible tool or object while avoiding surrounding clutter.}
    \label{fig:cover}
\end{figure}

On a platform or industrial workbench, the target may be partly hidden by nearby equipment, observed from a single robot stance, and reconstructed from noisy onboard depth measurements. A grasp that looks plausible on the visible surface can fail once the gripper body, approach direction, hidden geometry, and arm reach are considered~\cite{bohg2013data,jiang2021synergies,9197318,10611725,9981499}. The challenge extends beyond predicting a 6-DoF pose. The robot must connect a semantic instruction to instance-level geometry, reason from a partial view, reject grasps with poor clearance, and coordinate the base and arm so that the chosen grasp is executable.

Recent grasping methods have improved 6-DoF grasp generation from depth or point clouds~\cite{ten2017grasp,sundermeyer2021contact,fang2023anygrasp}. GPD and GPG sample antipodal candidates from observed point clouds~\cite{ten2017grasp,gpg2016}. Contact-GraspNet and AnyGrasp use learned representations to improve grasp prediction in clutter~\cite{sundermeyer2021contact,fang2023anygrasp}. Related works also explore instance-centric grasping, target-oriented cluttered grasping, and 3-D representations for manipulation~\cite{9197318,10611725,9981499,ji2024graspsplats,li2024robogsim}. These methods provide strong tools, but field manipulation still requires the full pipeline from an operator's object request to a grasp that the robot can execute with its own sensors, base, arm, and gripper.

Language-conditioned perception is a practical interface for this setting. Open-vocabulary detectors and promptable segmentation models, such as Grounding DINO~\cite{liu2024grounding} and SAM~2~\cite{ravisam}, allow an operator to specify a target with natural language instead of a fixed class label or object template~\cite{van2024open,11245995,ren2024grounded}. However, bridging semantic grounding and masks to reliable 3D grasp execution under partial observation remains challenging. The robot must convert grounded masks into object-centric geometry, infer missing surfaces, and generate grasps that remain feasible under real motion and safety constraints. Existing works typically address perception, grasp prediction, or execution individually rather than as a unified execution-aware pipeline.

\begin{figure*}[!t]
    \centering
    \includegraphics[width=\linewidth]{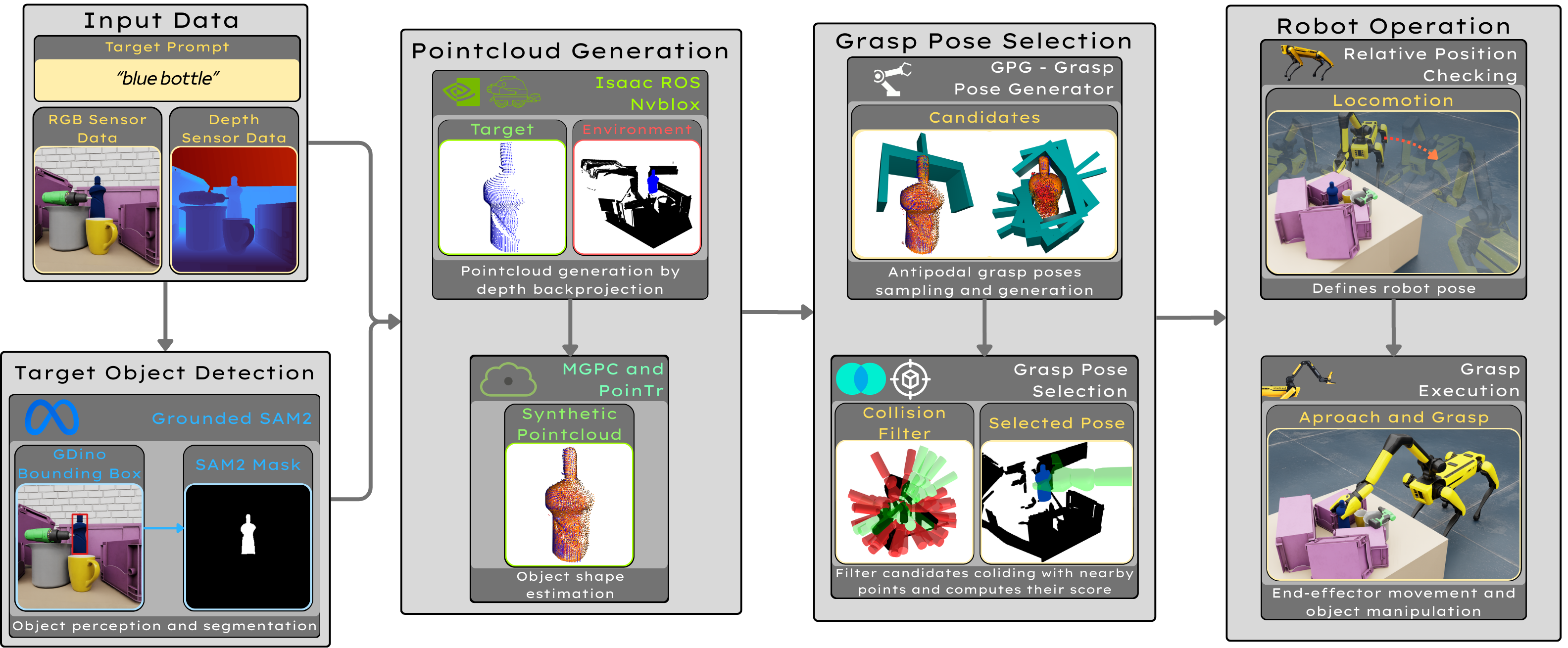}
    \caption{System overview of the proposed viewpoint-agnostic grasping pipeline. The system receives a natural-language target prompt (e.g., ``blue bottle'') together with synchronized RGB and depth observations. The prompt is grounded in RGB using Grounding DINO~\cite{liu2024grounding} to obtain a target bounding box and SAM~2~\cite{ravisam} to produce an instance mask. The mask is then used to extract an object-centric partial point cloud from depth via Isaac ROS Nvblox~\cite{millane2024nvblox} using depth backprojection. To mitigate occlusions and sparse depth, the object geometry is completed in two stages: MGPC~\cite{liu2026mgpc} generates synthetic points conditioned on the prompt, RGB, and the partial point cloud, and PoinTr~\cite{yu2021pointr} further densifies the geometry by completing fixed-size local patches. Given the densified point cloud, Grasp Pose Generator (GPG)~\cite{gpg2016} samples antipodal 6-DoF grasp candidates, which are collision-filtered against nearby scene points and ranked to select an execution-feasible grasp. Finally, the robot executes a state-machine locomanipulation routine that (when needed) repositions the base for reachability and clearance, followed by end-effector approach, grasp closure, and object lift.}
    \label{fig:diagram}
\end{figure*}

In this paper we present a language-guided grasping system for a legged manipulator under the single-view and cluttered conditions common in inspection and maintenance work. The key objective is reliability in unstructured scenes, such as offshore oil platforms, by developing a framework capable of identifying the intended object in clutter, estimating graspable geometry from incomplete views, and executing a collision-free, stable grasp while enforcing safety constraints throughout. To achieve this, we integrate open-vocabulary, language-guided target selection with object-centric 3D estimation and spatial-aware grasping in an integrated execution-aware pipeline. Target selection is grounded in RGB observations via VLM-guided querying and open-vocabulary detection and segmentation, enabling scalable specification of novel objects without task-specific retraining. Geometry is then derived from RGB-D, extracting and aggregating object-centric point clouds from masked observations, with back-projected depth compensation to improve robustness to missing returns. We explicitly address occlusion by using point cloud completion models, resulting in a grasping-ready representation even when significant portions of the object are unobserved. Finally, grasps are selected for safe execution by incorporating approach feasibility, clearance, and collision risk, and are executed through motion planning on the real robot. Overall, our approach enables safe, repeatable grasp acquisition in clutter at runtime from partial observations, providing a scalable pipeline from target selection to execution in real-world environments.

The main contributions of this paper are summarized as follows:
\begin{itemize}
    \item \textbf{Unified End-to-End Framework:} An integrated pipeline that bridges natural language-driven target specification and execution-feasible grasping for mobile legged robots operating in cluttered environments.
    
    \item \textbf{Execution-Aware Grasp Selection:} A grasping strategy that incorporates approach feasibility, clearance, collision constraints, and whole-body kinematic limits to ensure reliable real-world execution.
    
    \item \textbf{Occlusion-Resilient Geometry Estimation:} An object-centric 3D reconstruction process from masked RGB-D observations with depth back-projection and MGPC-based shape completion to handle severe partial observations.
    
    \item \textbf{Mobile Loco-manipulation:} Coordinated base repositioning and arm execution driven by execution-feasible grasp selection to improve accessibility and reliability in clutter.
    
    \item \textbf{Real-World Validation:} 
    Real-robot experiments on a legged mobile manipulation platform demonstrate the feasibility of language-guided grasp acquisition for supervised inspection and maintenance scenarios.

\end{itemize}

The rest of this paper is organized as follows: Section~\ref{sec:methods} describes the proposed view-agnostic grasping pipeline, and Section~\ref{sec:experiments} presents the experiments conducted to validate the framework. Section~\ref{sec:results} presents the results of this study, and finally, Section~\ref{sec:conclusion} concludes the paper.

\section{METHODS}\label{sec:methods}

The proposed pipeline for viewpoint-agnostic grasping is based on three main modules: A) object detection and segmentation, B) point cloud generation and estimation, C) grasp pose generation, selection, and D) execution. The framework, shown in Fig.~\ref{fig:diagram}, takes RGB-D images from the robot's cameras as input and is able to perform grasping and locomanipulation in cluttered environments. The current system is intended for supervised inspection and maintenance tasks in which the target object is static during planning and remains at least partially visible to the robot. The entire framework is implemented using Robot Operating System (ROS) 2~\cite{quigley2009ros}\footnote{The repository will be made available at Github}.

\subsection{Detection and Segmentation}
\label{sec:detseg}

The perception stage takes RGB images from the robot's front cameras as input and outputs an object-centric instance mask that is later used to extract 3D geometry from the volumetric map (Sec.~\ref{sec:pcgen}). While the robot provides stereo RGB-D data at 15~Hz, detection and segmentation are performed using RGB only. Depth is used to generate object geometry (Sec.~\ref{sec:pcgen}) via nvblox-based depth integration and point cloud extraction~\cite{millane2024nvblox}.

\paragraph{Target specification and open-vocabulary detection, segmentation, and tracking}

The operator specifies the target object via a natural-language command (e.g., ``blue bottle'') that should be within the robot's view range. Given the text query, the target is initially localized using the open-vocabulary detector GroundingDINO~\cite{liu2024grounding}, which returns candidate bounding boxes with confidence scores. We select the highest-scoring box associated with the query, denoted $B^\star$, and use it to initialize SAM~2 for instance segmentation and \emph{video tracking}~\cite{ravisam}. During execution, SAM~2 maintains the target mask across subsequent frames. GroundingDINO is invoked again only if tracking fails (i.e., SAM~2 outputs no valid mask), at which point detection is re-initialized and tracking resumes. If no valid hypothesis is produced after re-initialization, the system does not proceed to grasp planning and continues acquiring observations.

\paragraph{Promptable instance segmentation and tracking}
To obtain a pixel-accurate object mask in clutter, we refine $B^\star$ using SAM~2~\cite{ravisam} as the segmentation model. The selected box $B^\star$ is passed as a box prompt to SAM~2, producing a binary instance mask $M$ for the target. We apply a lightweight morphological erosion to $M$ using OpenCV to suppress boundary leakage into nearby clutter and improve mask robustness for 3D extraction. The output of this module is the mask $M$, which is then used to extract an object-centric point cloud from RGB-D observations (Sec.~\ref{sec:pcgen}).

\subsection{Point Cloud Generation and Estimation}
\label{sec:pcgen}

This stage converts the RGB-only instance mask into object-centric 3D geometry suitable for grasp synthesis under partial observations. We use the robot's RGB-D cameras to obtain depth images and leverage Isaac ROS nvblox for efficient GPU-accelerated depth processing and \emph{point cloud extraction}~\cite{millane2024nvblox}. In contrast to volumetric TSDF/ESDF mapping, our pipeline operates directly on point clouds and uses back-projected depth compensation to reduce sparsity and missing returns.

\paragraph{Object-centric point cloud extraction}
For each RGB-D frame, nvblox is employed to back-project the depth image and produce a registered scene point cloud in the robot frame. We then apply the instance mask $M$ (Sec.~II-A) to retain only points that project inside the segmented target region, resulting in an object-centric partial point cloud $P_{\text{partial}}$. To improve robustness under partial views, masked point clouds are aggregated over time in a common reference frame using the robot state estimation. This multi-frame accumulation increases surface coverage while remaining lightweight and scalable for real-time operation.

\paragraph{Back-projected depth compensation}
Depth in clutter often contains holes, flying pixels near discontinuities, and missing returns on thin or specular structures. To mitigate this, we apply a back-projected depth-compensation step in the nvblox point cloud generation pipeline before extracting $P_{\text{partial}}$. Small depth holes are filled, and outliers are attenuated using local neighborhood consistency in the image plane, producing a denser and more stable object-centric point cloud without requiring a volumetric distance map.

\paragraph{Completion from partial observations (MGPC)}
Even after multi-frame accumulation and depth compensation, $P_{\text{partial}}$ remains incomplete due to self-occlusions and backside surfaces. We apply MGPC~\cite{liu2026mgpc} to estimate missing geometry by leveraging multimodal context (prompt, RGB, and the partial point cloud). MGPC requires a fixed-size point input. Therefore, we subsample $P_{\text{partial}}$ to 2048 points and infer a synthetic point set $P_{\text{mgpc}}$ with 8192 points, the default model output. We then merge synthetic and observed points to obtain an intermediate cloud, as given by Eq.~\ref{eq:mgpcpartial}.
\begin{equation}\label{eq:mgpcpartial}
P_{\text{mid}} = P_{\text{partial}} \cup P_{\text{mgpc}},
\end{equation}

This increases surface coverage while remaining consistent with the visible geometry. In practice, for an input of $N$ observed points, this step results in $N + 8192$ points prior to PoinTr~\cite{yu2021pointr} refinement.

\paragraph{Point cloud refinement and densification (PoinTr)}
Since the Grasp Pose Generator (GPG) is sensitive to normal estimation quality, we further densify the object geometry using PoinTr~\cite{yu2021pointr}, a point-cloud-only completion model. Since PoinTr requires a fixed input size, we decompose $P_{\text{mid}}$ into overlapping local patches of 2048 points using a KD-tree neighborhood query. Each patch is completed independently by PoinTr, generating additional points that refine the local surface structure. The union of completed patches is then merged with $P_{\text{mid}}$ to obtain the final densified cloud $P_{\text{complete}}$, which typically increases the point count substantially (e.g., from $\sim$2k to $\sim$10k after MGPC, and further after PoinTr refinement). The final point cloud $P_{\text{complete}}$ is passed to the grasp module for 6-DoF grasp proposal generation, selection, and safe execution.

\begin{figure}[t!]
    \centering
    \hfill
    \begin{subfigure}[b]{0.20\textwidth}
        \centering
        \includegraphics[angle=-90, width=\linewidth, trim={0 0 0 0}, clip]{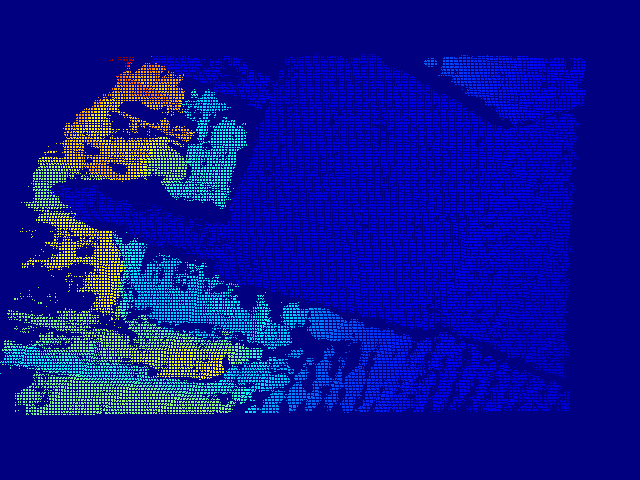}
        \caption{Stereo registered depth.}
        \label{fig:stereo1}
    \end{subfigure}
    \hfill 
    \begin{subfigure}[b]{0.20\textwidth}
        \centering
        \includegraphics[angle=-90, width=\linewidth, trim={0 0 0 0}, clip]{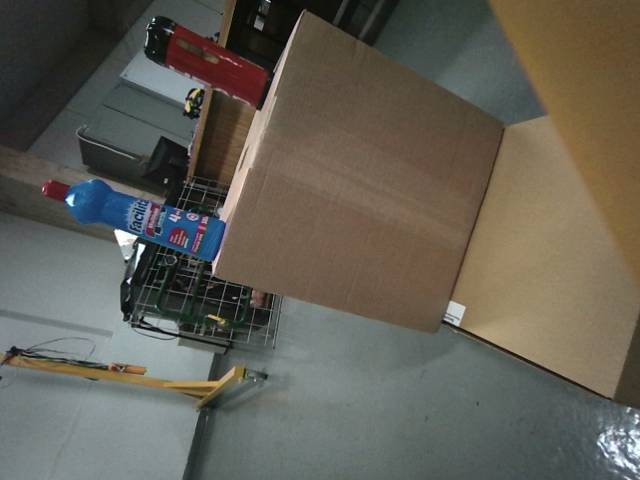}
        \caption{RGB Image.}
        \label{fig:rgb1}
    \end{subfigure}
    \hfill
    \caption{Spot front-left registered stereo and RGB images example taken with the robot still. The images showcase the noise and limited resolution of the available sensors.}
    \label{fig:spot_cameras1}
\end{figure}

\subsection{Grasp Pose Generation and Selection}
\label{sec:graspgen}

Given the completed object point cloud $P_{\text{complete}}$ (Sec.~\ref{sec:pcgen}), we sample 1000 candidate 6-DoF grasps and select a single grasp $g^\star$ for execution under clutter and reachability constraints. Since GPG relies on surface normal estimates, densifying $P_{\text{complete}}$ improves normal stability and increases the diversity of feasible grasp hypotheses. We use the Grasp Pose Generator (GPG) to sample a candidate set $\mathcal{G}=\{g_i\}$ on $P_{\text{complete}}$, where each candidate $g_i=(\mathbf{p}_i,\mathbf{R}_i)$ defines a gripper pose with position $\mathbf{p}_i$ and orientation $\mathbf{R}_i$ in the robot frame. We configure GPG's antipodal sampling parameters (jaw width and contact constraints) to match the Spot jaw gripper geometry, ensuring that sampled grasps are physically realizable on the hardware.

\paragraph{Collision filtering}
To enforce feasibility in cluttered environments, each candidate is validated by collision checking against the local scene geometry. We evaluate gripper-environment intersections using a parallelized kernel that tests the gripper mesh (at pose $g_i$) against the environment point cloud in a neighborhood around the target. Candidates that result in collisions with the scene are rejected, resulting in a filtered set $\mathcal{G}_{\text{free}}\subseteq\mathcal{G}$.

\paragraph{Heuristic ranking}
From $\mathcal{G}_{\text{free}}$, we select the grasp that best trades off approach feasibility and grasp stability using a weighted cost function, given by Eq.~\ref{eq:grasp_cost}.
\begin{equation}
\mathcal{C}(g_i) = 
w_\theta \, |\Delta\theta_i|
\;+\;
w_\phi \, \phi_i
\;+\;
w_c \, \|\mathbf{p}_i - \mathbf{c}\|
\;+\;
\mathcal{P}(r_i),
\label{eq:grasp_cost}
\end{equation}
where $\mathbf{c}$ is the centroid of $P_{\text{complete}}$ and $r_i=\|\mathbf{p}_i-\mathbf{p}_{\text{base}}\|$ is the distance from the robot base to the candidate grasp position.

\begin{itemize}
    \item \textbf{Alignment ($|\Delta\theta_i|$):} angular deviation (rad) between the base-to-target direction and the grasp approach direction. This biases grasps toward the robot's nominal approach and typically requires less base repositioning.
    \item \textbf{Approach bias ($\phi_i$):} binary penalty that discourages unfavorable approach directions (e.g., approaching from below), which are more likely to be kinematically constrained or blocked in clutter.
    \item \textbf{Centrality ($\|\mathbf{p}_i - \mathbf{c}\|$):} distance to the object centroid, which favors more centered grasps that are less sensitive to partial geometry and contact uncertainty.
    \item \textbf{Reach constraint ($\mathcal{P}(r_i)$):} hard penalty enforcing a maximum reach radius $r_{\max}$, as given by Eq.~\ref{eq:reach_penalty}.
\end{itemize}

\begin{equation}
\mathcal{P}(r_i)=
\begin{cases}
0, & r_i \le r_{\max},\\
M, & r_i > r_{\max},
\end{cases}
\label{eq:reach_penalty}
\end{equation}
where $M$ is a large constant.

The final grasp is chosen as given by Eq.~\ref{eq:grasp_select}.

\begin{equation}
g^\star = \arg\min_{g_i \in \mathcal{G}_{\text{free}}}\; \mathcal{C}(g_i).
\label{eq:grasp_select}
\end{equation}

In hazardous environments, the weights $(w_\theta,w_\phi,w_c)$ and limit $(r_{\max})$ provide explicit parameters to tune the grasp selection toward safer, more reliable grasps in cluttered scenes.

\subsection{Grasp Execution and Motion Control}
\label{sec:graspexe}

Execution of the selected grasp $g^\star$ is managed by a finite-state machine that coordinates base repositioning and arm motion to ensure reachability. If the grasp is not reachable from the current stance, the robot commands a base waypoint along the grasp approach direction at a stand-off distance that improves manipulability while keeping the target within the arm workspace. This repositioning step is used to satisfy reachability and clearance constraints prior to arm motion.

After base alignment, the manipulation sequence proceeds in two stages:

\paragraph{Pre-grasp approach}
We define a pre-grasp pose $g_{\text{pre}}$ by offsetting $g^\star$ along the gripper approach axis by a safety distance $\delta$:
\begin{equation}
g_{\text{pre}} = g^\star \oplus (\delta \, \hat{\mathbf{x}}),
\end{equation}
where $\hat{\mathbf{x}}$ is the local approach axis of the gripper frame and $\oplus$ denotes pose composition.

\paragraph{Final insertion and closure}
From $g_{\text{pre}}$, the end-effector executes a short Cartesian insertion of length $\ell$ (in our implementation, $\ell=\SI{5}{cm}$) along the approach axis to reach $g^\star$. Once the final pose is reached, the robot gripper service commands the gripper to close to secure the object.











\section{EXPERIMENTS}\label{sec:experiments}

We evaluate the proposed viewpoint-agnostic grasping pipeline on a Boston Dynamics Spot equipped with an arm, in cluttered scenes representative of unstructured deployments (See Fig.~\ref{fig:cover}). The goal of the experiments is to quantify the benefits of our geometry estimation and grasping generation pipeline under partial observations, in particular, point cloud completion and spatial-aware grasp selection, compared with a view-dependent baseline that fixes the robot position upon detection.

\subsection{End-User Requirements and Scope}
\label{sec:end_user_scope}

The intended end users are remote inspection and maintenance operators responsible for critical infrastructure such as offshore oil and gas facilities. The practical requirements considered here are to let a human operator specify the desired object semantically, use onboard RGB-D sensing without external motion capture or pre-scanned CAD models, handle clutter and partial observations around the target, and select grasps that respect base-arm reachability and approach clearance. The present experiments target early manipulation validation for a larger inspection-and-maintenance robotics project, before offshore deployment, dynamic scenes, rough terrain, certified safety, and continuous online replanning are introduced.

\begin{figure}[!t]
    \centering
    \includegraphics[width=\linewidth]{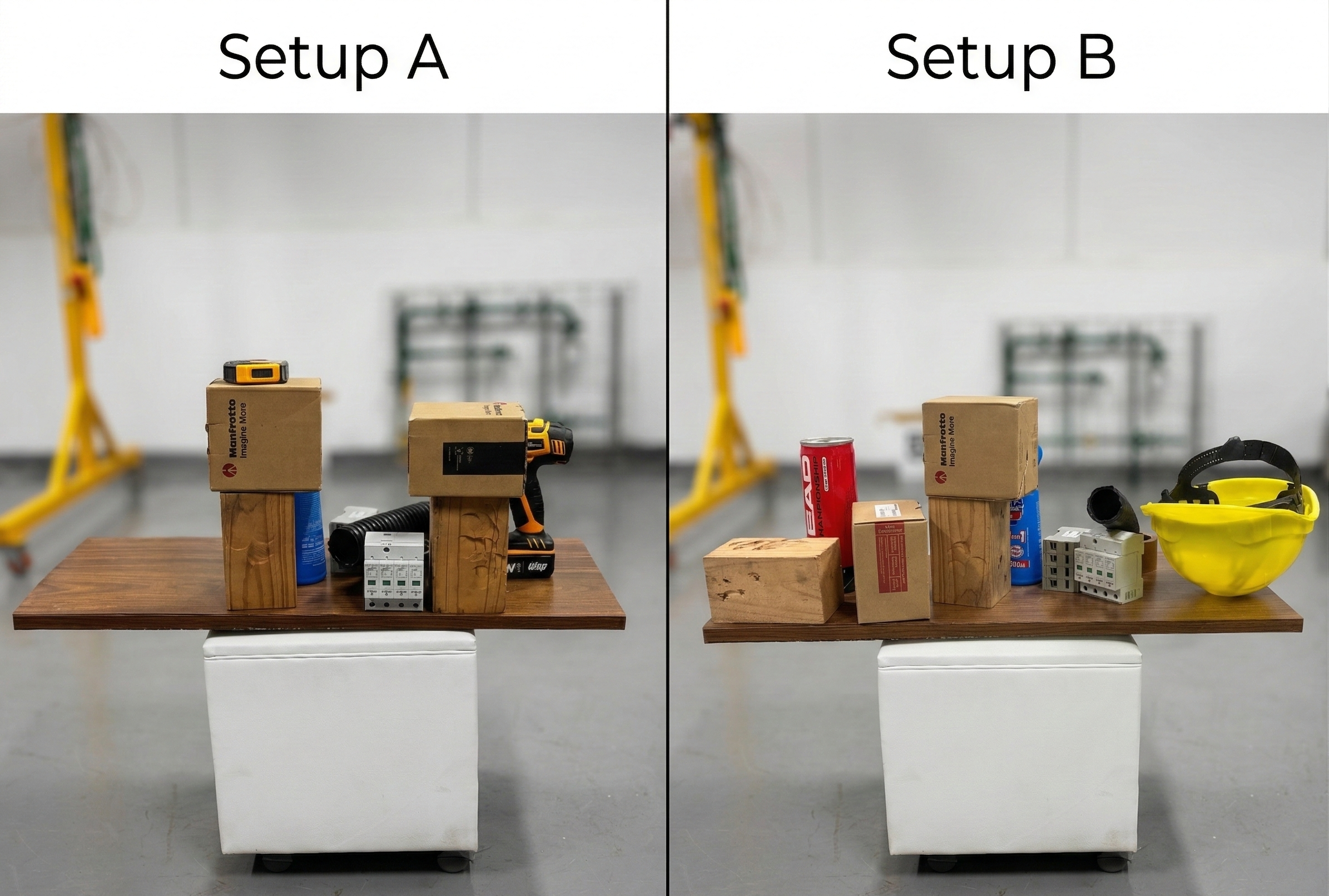}
    \caption{Experimental setups for evaluating the viewpoint-agnostic grasp pipeline. The environments consist of cluttered industrial and household objects. Setup A (left) was used for experiments to identify and grasp a power drill partially obscured by boxes and electrical components. Setup B (right) requires the pipeline to target a blue bottle situated behind different boxes. These configurations test the model's ability to generate grasps despite the challenging scenarios.}
    \label{fig:wardrobe_setup}
\end{figure}

\subsection{Hardware and Software}
\label{sec:hw}

All experiments were conducted on a Boston Dynamics Spot quadruped equipped with the Spot Arm (6-DoF) and a jaw gripper, using the robot's onboard RGB-D sensing. Computation was split between the robot and an external workstation connected via Ethernet. The workstation was used to run the perception models and geometry/grasp optimization, while Spot executed locomotion and manipulation through its onboard controllers.

\paragraph{Sensing}
Spot provides five RGB-D sensors. In the experiments, the front-left and front-right cameras were used. RGB images were streamed at \SI{15}{Hz} with VGA resolution ($640\times480$), and stereo depth images were streamed at $424\times240$ pixels. Examples of those images are shown in Fig.~\ref{fig:spot_cameras1}.


\paragraph{Compute}
The external workstation is equipped with two NVIDIA RTX A2000 GPUs (12~GB each), 125~GB system memory, and an Intel\textsuperscript{\textregistered} Xeon\textsuperscript{\textregistered} w3-2435 CPU ($\times$16 cores). All learning-based components (GroundingDINO, SAM~2, and MGPC) and the associated point-cloud processing were processed on this workstation.

\paragraph{Software and communication}
Robot communication and command execution are handled through the Boston Dynamics Spot SDK. The overall system is implemented in ROS~2 and deployed using Docker containers (ROS~2 Humble and Jazzy), with the workstation acting as the primary compute node and Spot as the execution platform. The pipeline runs end-to-end in real time, synchronized with the RGB-D camera stream and the state-machine execution described in Sec.~\ref{sec:graspexe}.

\begin{table*}[!t]
\centering
\renewcommand{\arraystretch}{1.2}
\setlength{\tabcolsep}{6pt}
\begin{tabular}{|c|c|c|>{\raggedright\arraybackslash}p{5.2cm}|>{\raggedright\arraybackslash}p{6.2cm}|}
\hline
\multicolumn{3}{|c|}{} &
\multicolumn{2}{c|}{\textbf{Success or Failure (alongside failure reason)}} \\ \hline
\textbf{Scenario} & \textbf{Run} & \textbf{Object} & \textbf{Our Method} & \textbf{Baseline (View-Dependent)} \\ \hline

\multirow{5}{*}{A} & 1 & drill & \cmark & \xmark\ (reachability failure) \\ \cline{2-5}
                   & 2 & drill & \cmark & \xmark\ (approach collision (clutter)) \\ \cline{2-5}
                   & 3 & drill & \cmark & \xmark\ (approach collision (clutter)) \\ \cline{2-5}
                   & 4 & drill & \xmark\ (reachability failure) & \xmark\ (reachability failure) \\ \cline{2-5}
                   & 5 & drill & \cmark & \xmark\ (approach collision (clutter)) \\ \hline

\multirow{5}{*}{B} & 1 & blue bottle & \cmark & \xmark\ (approach collision (target)) \\ \cline{2-5}
                   & 2 & blue bottle & \cmark & \cmark \\ \cline{2-5}
                   & 3 & blue bottle & \cmark & \xmark\ (approach collision (clutter)) \\ \cline{2-5}
                   & 4 & blue bottle & \cmark & \cmark \\ \cline{2-5}
                   & 5 & blue bottle & \cmark & \cmark \\ \hline

\multicolumn{3}{|r|}{\textbf{Total success rate}} & \textbf{(9/10)} & \textbf{(3/10)} \\ \hline
\end{tabular}
\caption{Success/failure outcomes from the experiments performed across scenarios A and B. The failure reason is presented whenever the robot fails to either reach or securely grasp the target object.}
\label{tab:success_failure}
\end{table*}

\paragraph{Implementation details}
For reproducibility, the full set of experimental parameters (perception thresholds, point-cloud completion settings, and grasp-selection weights) is documented in the code repository.


\subsection{Experimental Setups}
Two cluttered tabletop setups, shown in Fig.~\ref{fig:wardrobe_setup} were used, differing mainly in the grasp target:
\begin{itemize}
    \item \textbf{Setup~1 (Drill):} the target object is a handheld drill placed among different objects and occluders.
    \item \textbf{Setup~2 (Blue bottle):} the target object is a blue bottle placed among different objects and occluders.
\end{itemize}
In both setups, the table height was comparable to the robot base height, allowing consistent visibility from the front cameras. The table surface contained a variety of objects and obstacles intentionally arranged to occlude the target and induce partial observations. The robot always faced the table at the start of each trial.

\subsection{Methods Compared}
For each setup, we compare:
\begin{itemize}
    \item \textbf{Ours (viewpoint-agnostic):} full pipeline including mask-conditioned point cloud extraction, multi-frame accumulation, MGPC and PoinTr point cloud completion, and mobile grasp selection/execution (see Section~\ref{sec:methods}).
    \item \textbf{Baseline (view-dependent):} A conventional grasping pipeline that uses the same perception front-end and GPG-based grasp candidate generation, followed by identical collision filtering and heuristic ranking. However, grasp planning is performed directly on the single-view partial point cloud without multi-frame accumulation or point cloud completion, and the robot executes grasps from its initial stance without base repositioning.
\end{itemize}

This baseline isolates the impact of the proposed viewpoint-agnostic geometry estimation and mobile grasp execution strategy. In contrast to our method, the robot must commit to a grasp using only the partial object geometry visible from the initial viewpoint, which reflects a common deployment setting in many existing grasp pipelines.

\subsection{Trial Protocol}
Each setup consists of 10 trials. For each setup, 5 trials are performed with our method and 5 with the baseline. To ensure a fair comparison, we use a paired protocol: for a given initial robot pose, we execute one trial with our method and one trial with the baseline, starting from the same position. Across pairs, the initial robot position is changed (while maintaining the same scene configuration and the robot facing the setup) to test robustness to viewpoint-dependent occlusion and depth sparsity. After each grasp attempt, the robot returns to its initial pose before starting the next trial.

Each trial follows the same sequence:
\begin{enumerate}
    \item \textbf{Perception:} the operator provides a natural-language target query (``drill'' or ``blue bottle''). The system detects and segments the target and extracts an object-centric point cloud.
    \item \textbf{Geometry estimation:} our method applies depth compensation and point cloud completion using MGPC and PoinTr; the baseline uses only the partial point cloud from the initial view.
    \item \textbf{Grasp selection:} grasp candidates are generated and ranked, producing a selected grasp pose $g^\star$.
    \item \textbf{Execution:} the robot repositions its base if needed to satisfy reachability, executes a pre-grasp approach, performs a final Cartesian insertion, and closes the gripper.
\end{enumerate}

\subsection{Success Criteria}
A trial is labeled as \textbf{successful} if the robot (i) closes the gripper on the target object, (ii) lifts it from the table, and (iii) maintains a stable grasp for a short verification period without dropping or slipping. We report success rates per setup and method, and analyze failure modes qualitatively in Sec.~\ref{sec:results}.

\section{RESULTS AND DISCUSSION}\label{sec:results}

\begin{figure*}[t]
    \centering
    \newlength{\figH}
    \setlength{\figH}{3.5cm} 
    \begin{subfigure}{0.24\textwidth}
        \centering
        \includegraphics[width=\linewidth, height=\figH, keepaspectratio]{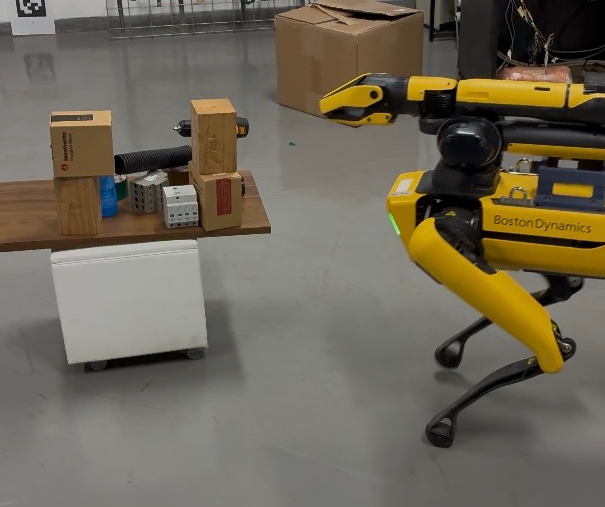}
        \caption{}
        \label{fig:frame1}
    \end{subfigure}
    \hfill
    \begin{subfigure}{0.24\textwidth}
        \centering
        \includegraphics[width=\linewidth, height=\figH, keepaspectratio]{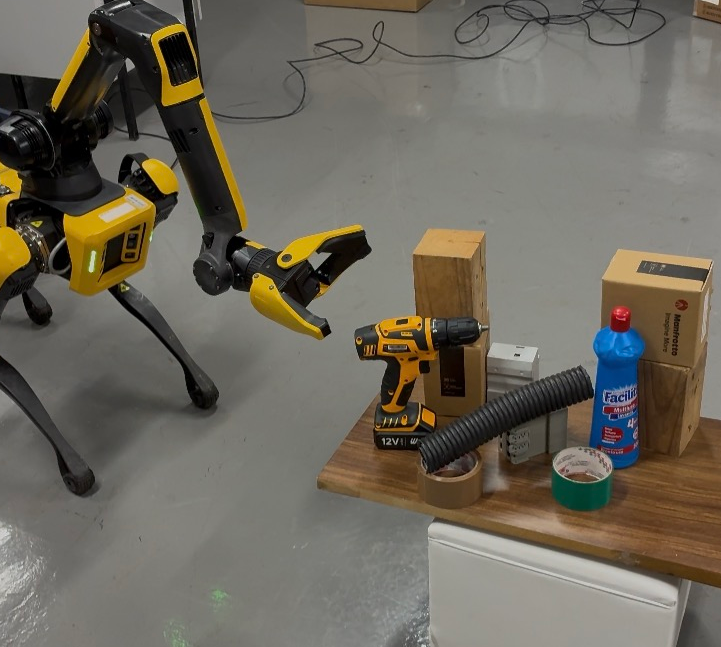}
        \caption{}
        \label{fig:frame2}
    \end{subfigure}
    \hfill
    \begin{subfigure}{0.24\textwidth}
        \centering
        \includegraphics[width=\linewidth, height=\figH, keepaspectratio]{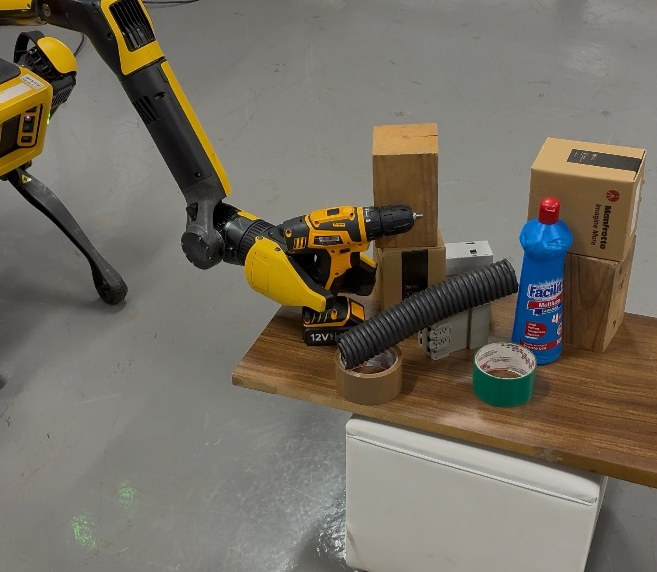}
        \caption{}
        \label{fig:frame3}
    \end{subfigure}
    \hfill
    \begin{subfigure}{0.24\textwidth}
        \centering
        \includegraphics[width=\linewidth, height=\figH, keepaspectratio]{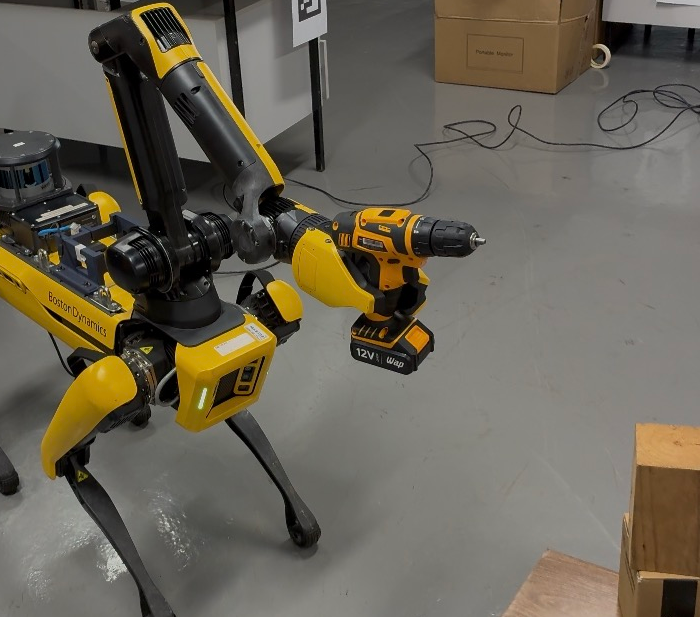}
        \caption{}
        \label{fig:frame4}
    \end{subfigure}
    \caption{Sequence demonstrating the grasp execution experiments using the proposed end-to-end pipeline on the real robot. (a) After language-guided target selection and instance segmentation, the system estimates object-centric 3D geometry from partial RGB-D observations (including completion), selects an execution-feasible 6-DoF grasp under collision and reachability constraints, and repositions the base to satisfy reachability and clearance for the planned approach. (b) The robot aligns to the target and commands the arm to a collision-free pre-grasp pose with a safety offset. (c) The end-effector executes a short Cartesian insertion along the grasp approach direction and closes the gripper to secure the object. (d) The object is lifted to confirm grasp success and stability under post-grasp interaction.}
    \label{fig:experiment_frames}
\end{figure*}

Table~\ref{tab:success_failure} summarizes the grasping outcomes across the two experimental scenarios, while Fig.~\ref{fig:experiment_frames} illustrates the experiments performed in this work. Our method achieved a total success rate of \textbf{9/10} trials, compared to \textbf{3/10} for the view-dependent baseline. In Scenario~A (drill), our pipeline succeeded in \textbf{4/5} trials, whereas the baseline failed in all \textbf{5/5} trials. In Scenario~B (blue bottle), our method succeeded in \textbf{5/5} trials, while the baseline succeeded in \textbf{3/5} trials. Overall, incorporating partial-observation geometry estimation and completion substantially improved robustness to clutter and occlusions in both setups. 



\subsection{Failure mode analysis}
From Table~\ref{tab:success_failure}, the observed failures modes (FM) that occurred in the experiments were divided into three categories:
\begin{itemize}
    \item \textbf{FM-1: Reachability failure}: the robot did not achieve a feasible pre-grasp configuration or could not reach the desired pose within the arm workspace.
    \item \textbf{FM-2: Approach collision (target)}: collision occurred while approaching the target, typically due to limited clearance along the approach direction.
    \item \textbf{FM-3: Approach collision (clutter)}: collision occurred with surrounding objects/occluders during approach.
\end{itemize}

The baseline fails predominantly due to approach collisions (FM-2/FM-3) in both scenarios, indicating that relying on the initial, view-dependent partial point cloud produces grasp candidates that are locally plausible but not executable once approach clearance is considered. In contrast, our method mitigates collision-driven failures, with only a single failure across all trials (Scenario~A, run~4) categorized as reachability failure (FM-1). This supports the hypothesis that improving object-centric geometry under partial observations (depth compensation + completion) increases the number of grasps that remain feasible under collision and reachability constraints.

\subsection{Limitations}
While the proposed pipeline improves grasp robustness in clutter, the current evaluation highlights practical limitations:

\paragraph{Target visibility and VLM grounding}
The target object must be sufficiently within the robot's field of view for the VLM-driven query, open-vocabulary detection, and segmentation to reliably identify the intended instance. If the prompt is ambiguous or the object is heavily occluded/out of view, grounding may fail or select the wrong instance, preventing the pipeline from proceeding.

\paragraph{Limited by VLM semantics}
Object specification is constrained by what the VLM and the open-vocabulary detector can reliably interpret from the scene. For domain-specific objects, unusual tools, or safety-critical part-level intent, fine-tuning (or task-specific prompt engineering and verification) may be required to maintain reliability.

\paragraph{Depth quality remains a challenge}
Even with depth compensation and point cloud completion, severe depth noise and limited sensor resolution can still degrade geometry estimation and collision checking. In our hardware setup, the stereo depth stream has limited resolution and exhibits noise, which can lead to missing surfaces and outliers in clutter. In such cases, completion may not sufficiently recover the correct shape for safe grasp selection, especially for thin or reflective objects.

\subsection{Key Observations}

The results highlight three important observations about the proposed pipeline. First, incorporating object-centric geometry estimation with partial-observation completion increases the reliability of grasp candidate generation under occlusions, enabling successful grasps even when only limited surface information is initially visible. Second, execution-aware grasp selection combined with collision-aware filtering produces grasps that remain feasible during real robot execution, reducing approach collisions that commonly occur when grasp planning relies solely on view-dependent geometry. Third, integrating mobile base repositioning with grasp planning improves accessibility in cluttered environments by allowing the robot to satisfy reachability and approach constraints before arm execution. Together, these components enable the robot to identify and execute grasps that remain feasible under real-world constraints, resulting in a substantial improvement in the success rate across both experimental scenarios.

\subsection{Implications for Inspection and Maintenance}
The deployable behavior demonstrated in this work is the complete pipeline from operator language to a physical grasp. The robot detects the requested object, builds local geometry, chooses a grasp that respects reach and clearance, and repositions when the arm alone is not enough. That behavior addresses a common gap in inspection robotics, where platforms can often observe assets but still struggle with simple physical handling tasks.

\section{CONCLUSION}\label{sec:conclusion}

This paper presented a language-guided grasping pipeline for legged mobile manipulation in field inspection and maintenance. The proposed system grounds a natural-language target in RGB using open-vocabulary detection and promptable segmentation, extracts object-centric geometry from RGB-D, and mitigates partial observations through depth compensation and point cloud completion. Based on the resulting geometry, the system generates and ranks 6-DoF grasp candidates using execution-aware, safety-oriented heuristics and executes the selected grasp via motion planning and a state-machine controller on a real Spot platform.

Experiments on two cluttered tabletop scenarios demonstrated that incorporating partial-observation geometry estimation and completion substantially improves grasp reliability compared to a view-dependent baseline, particularly in the presence of occlusions and limited clearance. These results support the main hypothesis that robust grasping in unstructured environments benefits from explicitly bridging semantic target grounding to object-centric 3D estimation and to execution-feasible grasp selection.  Future work will expand the object set and trial count, add external grasping baselines, separate the effects of completion and repositioning, and verify the completed geometry online during base and arm motion.



\bibliographystyle{IEEEtran}
\bibliography{IEEEabrv,references}

\end{document}